# HAVANA: Hard negAtiVe sAmples aware self-supervised coNtrastive leArning for Airborne laser scanning point clouds semantic segmentation

Yunsheng Zhang, Jianguo Yao, Ruixiang Zhang, Siyang Chen, Haifeng Li[*]

School of Geosciences and Info-Physics, Central South University, Changsha, Hunan, 410083, China

(zhangys@csu.edu.cn; 2606550285@qq.com; 2292225098@qq.com; 121555602@qq.com; lihaifeng@csu.edu.cn)

***Abstract:*** Deep Neural Network (DNN) based point cloud semantic segmentation has presented significant achievements on large-scale labeled aerial laser point cloud datasets. However, annotating such large-scaled point clouds is time-consuming. Due to density variations and spatial heterogeneity of the Airborne Laser Scanning (ALS) point clouds, DNNs lack generalization capability and thus lead to unpromising semantic segmentation, as the DNN trained in one region underperform when directly utilized in other regions. However, Self-Supervised Learning (SSL) is a promising way to solve this problem by pre-training a DNN model utilizing unlabeled samples followed by a fine-tuned downstream task involving very limited labels. Hence, this work proposes a hard-negative sample aware self-supervised contrastive learning method to pre-train the model for semantic segmentation. The traditional contrastive learning for point clouds selects the hardest negative samples by solely relying on the distance between the embedded features derived from the learning process, potentially evolving some negative samples from the same classes to reduce the contrastive learning effectiveness. Therefore, we design an AbsPAN (Absolute Positive And Negative samples) strategy based on k-means clustering to filter the possible false-negative samples. Experiments on two typical ALS benchmark datasets demonstrate that the proposed method is more appealing than supervised training schemes without pre-training. Especially when the labels are severely inadequate (10% of the ISPRS training set), the results obtained by the proposed HAVANA method still exceed 94% of the supervised paradigm performance with full training set. Furthermore, we found that our HAVANA has a significant effect on preserving geometry structure, and more importantly, such a method surpasses supervised learning methods in dense urban laser point clouds with many geometric instances.

***Keywords:*** ALS point cloud, semantic segmentation, self-supervision, end-to-end

## 1. Introduction

Airborne Laser Scanning (ALS) point clouds provide a compact and efficient representation of real-world 3D scenes, and have become a standard spatial-temporal geographic data source (Liu, 2019). However, the original point clouds only contain spatial coordinates or other auxiliary information, prohibiting a computer from comprehending a scene and obtaining structural information for subsequent applications. Thus, semantic segmentation assigns a label to each point in the point cloud, which is a typical pre-task for complex 3D model reconstruction (Zhu et al., 2018; Guo et al., 2020; Wang et al., 2020) and digital elevation model generation (Hu and Yuan, 2016).

In the early stages, semantic segmentation commonly relied on supervised learning methods using hand-crafted features (Weinmann et al., 2015a), and these methods can be divided into two categories. The first category is point-based methods, which are summarized by Weinmann et al. (2015b) into four steps: neighborhood selection, feature extraction, feature selection, and supervised classification, with the feature extraction stage having a significant impact on the final classification result. The second category includes statistical contextual-based methods (Niemeyer et al., 2014; Shapovalov et al., 2010; Xiong et al., 2011) that exploit each point's context characteristics and outperform point-based methods. Supervised paradigm machine learning based on hand-crafted features (Rusu et al., 2008; Tombari et al., 2010), and classical classification algorithms, such as random forest (Sun and Lai, 2014) and SVM (Weinmann et al., 2013), to complete supervised semantic segmentation of ALS point clouds. However, these traditional handcrafted-feature based methods limit point cloud semantic understanding, as they heavily depend on low-level features, have low accuracy, and have poor transferring ability.

Along with the development of deep learning methods, researches began to apply the deep learning methods to point cloud semantic segmentation. The first attempt was to remap the point cloud (3D data) to image representation (2D data), on which image semantic segmentation was performed, and the results were remapped into point clouds (2D to 3D remapping) (Kalogerakis et al., 2017; Zhao et al., 2018; Boulch et al., 2017). Aside from 2D CNN (Convolutional Neural Networks), some works tried to directly semantically segment the point clouds utilizing 3D CNN solutions (Maturana and Scherer, 2015; Schmohl and Sorgel, 2019; Tchapmi et al., 2017). Unlike these methods involving regular CNN, Qi et al. (2016) developed PointNet, a pioneering point-based semantic segmentation network that directly exploited irregular point clouds. Researchers also improved PointNet to PointNet++ (Qi et al. 2017), boosting direct point cloud segmentation research, such as

PointSIFT (Jiang et al., 2018), PointCNN+A-XCRF (Arief et al. 2019), and GACNN (global-local graph attention convolution neural network) (Cwab et al. 2021). It is very convenient for advanced ALS systems to capture a large number of point clouds, but it imposes a significant burden on training a classifier based on deep learning methods. Therefore, researchers focused on lightweight and efficient network architectures, such as RandLA-Net (Hu et al., 2020) and KPConv (Thomas et al., 2019), which achieved significant performance on point cloud semantic segmentation tasks. Nevertheless, these methods require many semantic annotations as prior information, with point cloud labeling being time-consuming and challenging.To the best of our knowledge, there is no point cloud dataset comparable to ImageNet.

The supervised learning paradigm based on the DNN method has obtained promising performance on point cloud semantic segmentation. This strategy requires many labeled points to increase the model's transferability and robustness, as the classification quality heavily relies on high-quality and complete point cloud datasets. For 2D images, researchers avoided over-reliance on semantic labels and solved the problem of the samples' insufficient semantic annotation by proposing the few-shot learning method. Existing work generates more samples (Shorten and Khoshgoftaar, 2019; Bowles et al. 2018), whereas other approaches focus on semi-supervised learning(Ahn and Kwak, 2018; Li et al., 2021). Several recent studies have used the self-supervised learning (SSL) paradigm to learn rich and diverse knowledge with fewer semantic labels (Araslanov and Roth, 2021; Wang et al. 2020). For example, Araslanov and Roth (2021) proposed the simple self-supervised framework SAC for domain adaptive semantic segmentation, which achieved promising results without the use of complicated methods. Wang et al. (2020) propose the self-supervised equivalent attention mechanism SEAM for pixel-level semantic segmentation, which outperformed using the same level of supervision on multiple data sets. In terms of the self-supervised learning paradigm in 2D images, self-supervised learning (SSL) is a viable alternative to solve the heavy reliance on manual labeling, as well as the future direction for currently unsolved problems (Ayush et al., 2020).

Several self-supervised learning methods have been proposed for 3D data recent years. Under the self-supervised mode, the auxiliary task exploited useful information from relevant tasks and learned weight to guide the point cloud semantic segmentation task, with a stronger inductive bias would be applied to concerned tasks. The SSL pattern has a greater potential in real-world semantic understanding applications (Sharma and Kaul, 2020; Liu et al., 2020; Hou et al., 2020; Rao et al., 2020; Xie et al., 2020; Sauder and Sievers, 2019; Poursaeed et al., 2020). Sauder and Sievers (2019) performed feature learning by restoring the point cloud's voxel positions and focusing on verifying the model's reconstruction reliability. Similarly, Poursaeed et al. (2020) proposed a method to predict rotations as the auxiliary task target. However, the corresponding learned features were only validated on shape classification and key-point detection tasks. Sharma and Kaul (2020) used a cover tree to encode point cloud hierarchical partitioning, where the proposed method generated variable-sized subsets with class labels. Liu et al. (2020) built contrastive learning for multi-modal RGB-D scans, while Rao et al. (2020) proposed a bidirectional reasoning scheme between the local structures and the global shape, taking unsupervised learning representation from data structures into account. Some works (Xie et al., 2020; Hou et al., 2020) used contrastive learning by constructing a positive and negative sample mining strategy. Nevertheless, these methods lack strong feature learning ability, and the traditional hardest negative mining strategy applied on point clouds encounters negative sample impure problem (Hou et al., 2020), resulting in the quantity of negative points adding very little to the ability of the model learning. As a result, this paper concentrated on SSL for semantic segmentation of point clouds.

To alleviate the reliance on point cloud annotation data, a method based self-supervised learning strategy for point clouds semantic segmentation, named HAVANA, which employed a mass of unlabeled points to pre-train a network for subsequent point cloud semantic segmentation. Specially, we design a k-means clustering guided negative samples selection method for the effective of the contrastive learning. Our main contributions are summarized below:

(1) A self-supervised contrastive learning scheme is introduced for semantic segmentation of point cloud. The meaningful information representation of unlabeled large-scale ALS point clouds is learned using contrastive learning, and then transferred to small local samples for high-level semantic segmentation tasks, resulting in improved semantic segmentation performance.

(2) We design AbsPAN, a strategy for selecting positive and negative samples for contrastive learning. This strategy employs an unsupervised clustering algorithm to remove potentially false-negative samples, ensuring that contrastive learning obtains meaningful information.

(3) Our results show that self-supervised learning is a multi-task learning scheme in which a useful pattern improve the generalization, outperforming the supervised learning model in terms of geometric structure preservation.

The remainder of the article is organized as follows: Section 2 describes the proposed method and Section 3 illustrates the experiments and analyzes the corresponding results. And finally, Section 4 concludes this work and give some outlooks.

## 2. Method

### 2.1. Overview

The proposed self-supervised learning paradigm for point clouds is illustrated in Fig. 2. Our technique aims to utilize unlabeled ALS point clouds to learn spatial invariance by designing self-supervised signals and transferring them to semantic segmentation tasks to reduce the reliance on semantic labels. The proposed method is composed of two parts. In the Self-supervised Pre-training part, contrastive learning is performed on large-scaled unlabeled point cloud by employing KP-FCNN (kernel point fully convolutional network) as the backbone network. In the Supervised Fine-tuning part, the pre-trained weight initiates the downstream semantic segmentation network training and fine-tune weight on a few labeled point clouds.

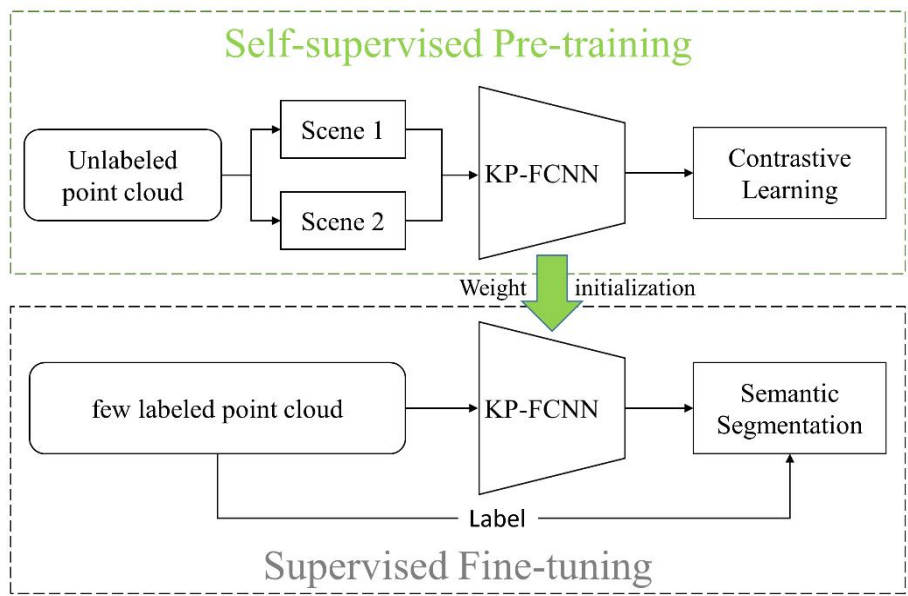

Fig. 2. The self-supervised learning paradigm for point cloud semantic segmentation

**2.2. Point cloud contrastive learning**

Self-supervised learning can be implemented by utilizing variety of specially designed supervised signals, with contrastive learning being selected due to its high representation learning ability. Contrastive learning is an unsupervised approach whose goal is to learn a mapping relationship $f$ from the original feature to the embedding space, so that the distance function is used to pull positive samples closer together and negative samples apart (Le-Khac et al., 2020; Chen et al., 2020). Our distance metric function focus on triplets with hard-negative mining (Wang et al., 2014). In particular, as shown in Eq. (1), for any anchors $x$, the goal of the contrastive learning is to learn an embedding function $f$ that meets the following condition:

$$\mathrm{L}(f(x), f(x^+)) \gg \mathrm{L}(f(x), f(x^-)) \qquad (1)$$

where $x^+$ are positive samples, and it denotes points similar or congruent to anchors $x$. $x^-$ are negative samples, and it denotes points dissimilar to anchors $x$. $\mathrm{L}$ is distance metric function used to measure the similarity between features.

The developed contrastive learning framework is illustrated in Fig. 3. The auxiliary pre-training task is described in section 2.2.1. For the point cloud feature embedding, KP-FCNN is selected as the backbone network and described in section 2.2.2. The key point of contrastive learning is to construct positive and negative samples. To ensure the purity of positive and negative samples, a negative samples mining method based on k-means clustering named AbsPAN ("Absolute Positive And Negative samples") is proposed and described in details in section 2.2.3. Our method optimizes a contrastive loss $\mathrm{L}_c$ on the extracted high-level features $V$ for a set of random geometrically transformed point cloud, it is completed by using KP-FCNN. The designed contrastive loss $\mathrm{L}_c$ is described in section 2.2.4.

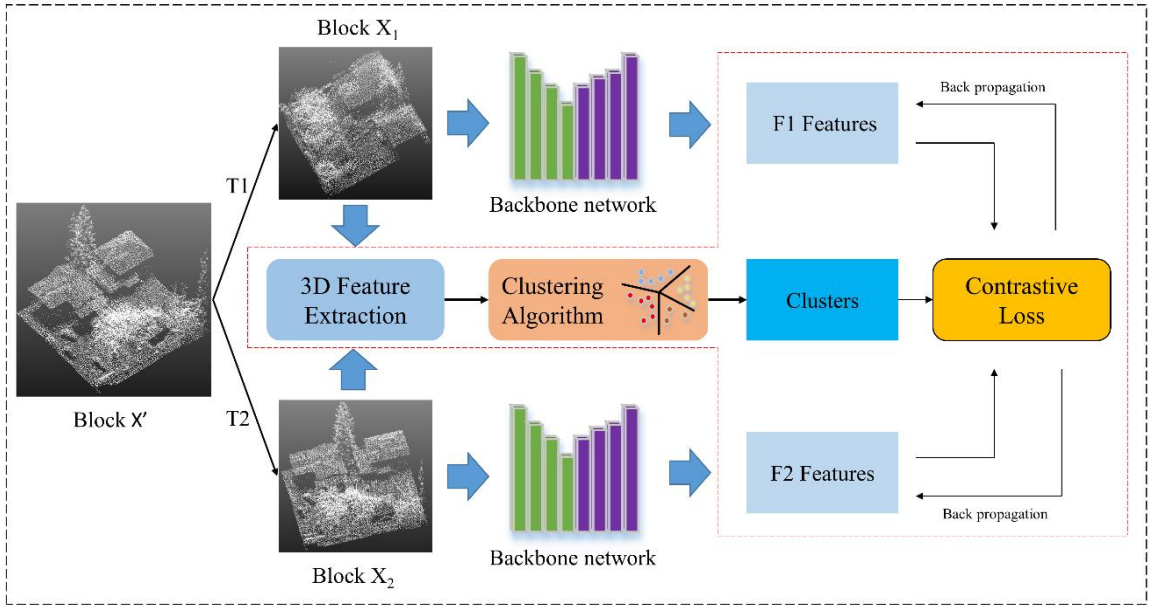

Fig. 3. Illustration of contrastive learning. For the backbone network, the green bars are encoders and the purple bars are decoders. The circled part of the red line represents our "AbsPAN" strategy selecting both positive and negative samples and optimize the loss function.

**2.2.1. Auxiliary pre-task**

For 3D ALS point clouds, the collected data is frequently excessive and thus difficult to be annotated. Therefore, the primary strategy for our auxiliary task would be to utilize these unlabeled point clouds. Inspired by work (Xie et al., 2020), the auxiliary tasks achieve point equivariance concerning a set of random similarity transformations (composed of rotation transformation and scaling, all the similarity transformations hereafter is referred to the rotation transformation and scaling). Concretely, similar to other metric learning algorithms (Song et al., 2016; Sohn, 2016; Choy et al., 2019b), we define a pair of matching points as a positive pair if they refer to the same point, otherwise, they form a negative pair. As illustrated in Fig. 3, given a point cloud dataset $X = \{x^1, \cdots, x^i, \cdots, x^N\}$, where N is number of all points. A spherical subset $X'$ is selected from X by using the random picking strategy using in KP-FCNN, and then two random similarity transformations are applied to $X'$ to obtain pair block $X_1 = \{x_1^1, \cdots, x_1^i, \cdots, x_1^n\}$, $X_2 = \{x_2^1, \cdots, x_2^j, \cdots, x_2^n\}$. For each point $x_1^i$ in block $X_1$, there is a corresponding point $x_2^j$ in block $X_2$, $(x_1^i, x_2^j)$ are matching points across two blocks. The unsupervised contrastive learning aims at extracting embedding features $v_s = f(X_s), s = \{1, 2\}$, where the embedding function $f$ maps the block to the feature space through the KP-FCNN framework, and $X_1$ is mapped to $V_1 = \{v_1^1, \cdots, v_1^i, \cdots, v_1^n\}$, $X_2$ is mapped to $V_2 = \{v_2^1, \cdots, v_2^j, \cdots, v_2^n\}$. The matched (positive) point features $(v_1^i, v_2^j)$ must be similar to each other and different from un-matched (negative) point features $v_2^k$. Contrastive learning accomplishes this goal by minimizing the contrastive loss function $\mathcal{L}_c$ (see Sec 2.2.4).

### 2.2.2. Backbone network

Due to the large amount of data utilized for pre-training in the proposed contrastive learning strategy, we adopt the KP-FCNN (Thomas et al. 2019) as the backbone network. Because the KP-FCNN stores the data to be processed locally at each layer, this pre-processing step is performed prior to training. Such a strategy increases training speed. Considering feature learning ability, the KP-FCNN utilizes strong deformable kernel convolution, which leads to improve object geometry awareness and information aggregation of adjacent categories without confusing them.

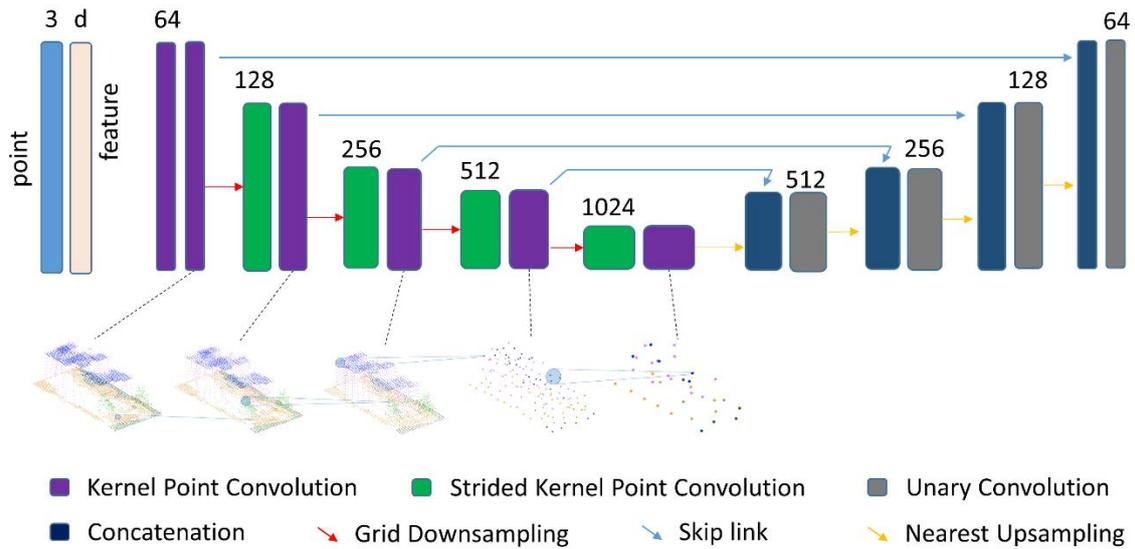

Fig. 4. Illustration of backbone network. The encoder (the first five layers) contains two convolutional blocks, kernel point convolution (KPConv) and strided kernel point convolution. The decoder (the last four layers) uses nearest upsampling, ensuring that the pointwise feature is added up. The features transferred from the encoder layer are concatenated to the upsampled ones by skip links. The concatenated features are processed by a unary convolution, which is the equivalent of a shared multi-layer perceptron (MLP) in PointNet.

KP-FCNN is quite flexible, allowing any number of kernel points and kernel point layouts, they are critical to the convolution operator, and can be set by the user to the balance computational cost and descriptive power. The architecture of KP-FCNN is illustrated in Fig. 4. This network uses a kernel point convolution to compute features and operates directly on point clouds. In practice, the input of KP-FCNN is a scene block $X_s$. In the five layers encoder, regular grid sampling is utilized as the downsampling method. In the four layers decoder, the upsampling method utilizes nearest sampling. At the end of the forward pass, a feature vector $V_s$, each have 64 dimensions is got.

It is worth noting that self-supervised learning network weights are used for initializing semantic segmentation weights. Slightly different from the pre-training backbone network, a layer of unary convolution is added to the last layer and then realize the conversion of 64 dimension to $D_{out}$ dimension in the downstream semantic segmentation task, $D_{out}$ is number of categories. We chose KP-FCNN as feature extractor, which can be replaced by other deep learning frameworks.

### 2.2.3. AbsPAN: Negative samples mining based on clustering

The PointContrast uses the hardest-negative mining scheme (referred as hardest-negative mining method hear-after) for subsequent point cloud semantic segmentation (Xie et al., 2020). However, changing the number of negatives has no discernible effect on segmentation result (Xie et al., 2020; Hou et al., 2020). This phenomenon contradicts our expectations. This could be because that the hardest negative pairs belong to the same semantic category. The hardest-negative mining methods treated each point independently, and leads to omission of the relationships between points. For point cloud semantic segmentation tasks, we believe that the ideal embedding space of auxiliary pre-task should accomplish intra-class compactness and inter-class dispersion. In the field of image, the idea of encoding semantic structure by clustering into embedding space has been proven effective (Tian et al., 2017; Li et al., 2020; Wang et al., 2021). To addressing the issue with the hardest-negative mining method, we propose a negative samples mining scheme named "AbsPAN" (Absolute Positive And Negative samples). The AbsPAN is defined as "a set of representative embeddings of semantically similar instances", with a focus on identifying "distinguishing negatives".

Fig. 5 illustrated the proposed schematic diagram of the negative samples mining processing. To alleviate the problem that the hardest negative pairs may belong to the same category. The transformed block 1 and 2 are clustered via an unsupervised clustering scheme based on k-means clustering. The clustering processing is performed on traditional handcrafted features are listed in Table 1. Because other information, such as intensity is often unreliable, these features are calculated solely using x-y-z coordinates. The used geometric are calculated based on covariance tensor $\Sigma_i$, as shown in Eq. (2).

$$\Sigma_i = \frac{1}{N} \sum_{n \in C_i^N} (c_n - \bar{c})(c_n - \bar{c}) \tag{2}$$

Where $C_i^N$ is the $N$ nearest point set to $C_i$, $\bar{c}$ is the medoid in $C_i^N$. In Table 1, the eigenvalues $\lambda_1 \geq \lambda_2 \geq \lambda_3 \geq 0$ and corresponding eigenvectors $e_1, e_2, e_3$ can be composed to describe geometric properties. Four hand-crafted features, as surface variation, verticality, normal vector $N_z$ and planarity selected for the clustering by referring to the relevance metric in work (Weinmann et al., 2015a). Surface variation is good at distinguishing between planar structure and non-planar planar structure, verticality is a strong feature to distinguish planes in different angles. These features are compact and robust, and are profitable for k-means clustering. The details can be found in work (Weinmann et al., 2015a).

Table1. Geometric features of the 3D point cloud.

| Geometric feature | Design formulas |
|---|---|
| Planarity | $(\lambda_2 - \lambda_3)/\lambda_1$ |
| Surface Variation | $\lambda_3/(\lambda_1 + \lambda_2 + \lambda_3)$ |
| Verticality | $1 - |e_3[2]|$ |
| Normal Vector | $N_x, N_y, N_z$ |

Based on clustering, a pseudo label is assigned to each point as shown in Fig. 5(b). After that, sample pairs is selected as follows:

1) $N_1$ ($N_1$ is experimentally set to 4096 in this paper) anchor point $x_1^i$ is randomly selected in block 1. Then correspondence matched point $x_2^j$ is chose as the positive sample for each anchor point $x_1^i$.

2) $N_2$ ($N_2$ is experimentally set to 2048 in this paper) anchor points are randomly selected from the pair $(x_1^i, x_2^j)$ for hardest negative sample selection. To ensure that the negative sample belongs to a different category, the point with the closest features (the feature is embedded by the KP-FCNN) to point $x_1^i$ in block 2 is chose as the hardest negative sample candidate and denoted as $x_2^{k-}$. If the pesudo label of $x_2^{k-}$ is not equal to $x_2^j$, $x_2^{k-}$ is a true candidate for the hardest negative sample of $x_1^i$. Otherwise, $x_2^{k-}$ will be removed and the next point with nearest feature will be validated until the true candidate is obtained. After getting the hardest negative sample of $x_1^i$ in block 2, the same process will be performed for the $x_2^j$ to search the hardest negative sample $x_1^{k-}$ in block 1.

Take Fig.5(c) and (d) as an example, if only the embedded feature distance is used as in the hardest negative mining algorithm, points $x_1^2$ and $x_2^2$ in Fig. 5(c) are regarded as a hardest negative pair. However, these points belong to the same category would confuse contrastive learning. Based on the pseudo label, points $x_1^2$ and $x_2^3$ will be selected as the true hardest negative pair (see Fig. 5(d)), because they are not in the same cluster (see Fig. 5(b) with different colors).

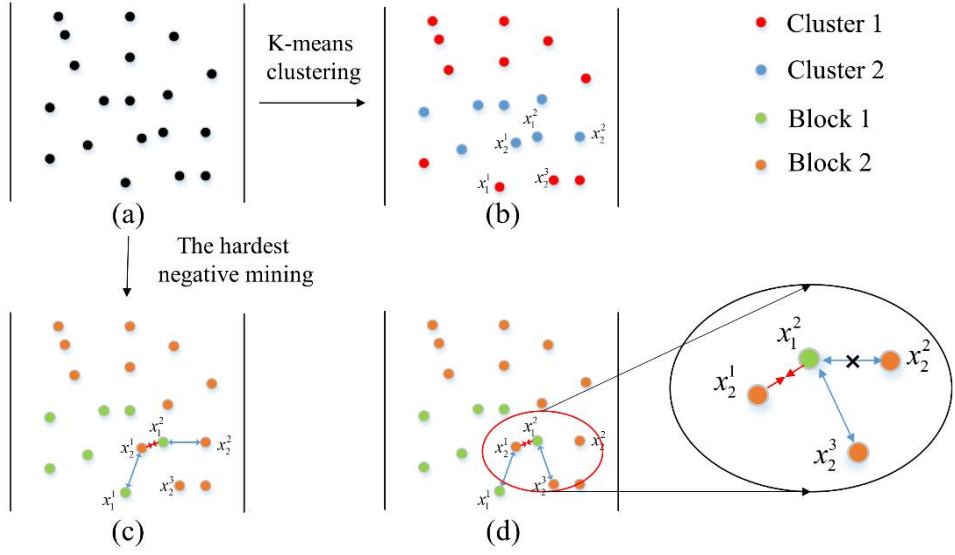

Fig. 5. Illustration of negative samples mining processing: (a) raw point clouds, (b) clustering results, (c) hardest negative candidates, (d) AbsPAN results. Among them, orange points represent block 1, green points represent block 2, (c) is in the embedding feature space of network, (b) is the cluster result by k-means clustering in geometric feature space, red points represent cluster 1, blue points represent cluster 2.

The proposed AbsPAN strategy benefits in two aspects: (1) in the hardest-negative mining, the hardest negative sample is the closest point in the extracted feature space to the positive sample. In practice, for anchor point $x_1^i$, we both have a correspondent matched point $x_2^j$ and a hardest negative $x_2^-$. Naturally, there is contradiction between pulling closer $x_1^i, x_2^j$, and pulling apart $x_1^i, x_2^-$, because $x_2^j$ and $x_2^-$ may have some of the same information, such as the same semantic structure. The positive pairs must have exactly the same information, while the negative pairs must be completely different. Negative pairs with similar geometric features will be removed. As a result which the positive and negative sample pairs are clearly distinguished. Clustering offers an useful factor for distinguishing between positive pairs and negative pairs. (2) The supervised signals consider geometric feature similarity, ensuring that more meaningful information are utilized in contrastive learning.

Regarding the downstream ALS semantic segmentation task, the richer migration information can improve the semantic segmentation result (Wang et al., 2020; Araslanov and Roth, 2021). Clustering and semantic segmentation have certain mutual information, our AbsPAN strategy can also be understood as "segmentation-aware hard-negative mining". The AbsPAN provides information relevant to downstream semantic segmentation tasks, and may help with point cloud semantic segmentation.

### 2.2.4. Loss function design

After obtaining the negative samples, it is necessary to develop a proper loss function to mine the contrastive semantic information of four types of points $(x_1^i, x_2^j, x_1^-, x_2^-)$. Inspired by the Hardest-Contrastive loss function (Choy et al., 2019b), the loss used function for the auxiliary task is defined as:

$$L_c = \sum_{(i,j)\in\theta} \left\{ \frac{\left[d(v_1^i, v_2^j) - t_p\right]_+^2}{|\theta|} + \frac{0.5\left[t_n - Q(v_1^i, v_2^k, c^{i,k})\right]_+^2}{|\theta_i|} + \frac{0.5\left[t_n - Q(v_1^k, v_2^j, c^{k,j})\right]_+^2}{|\theta_j|} \right\} \quad (3)$$

Where the first part of Eq. (3) is used to pull closer positive pairs in the learned feature space, the latter two parts of Eq. (3) are used to push apart negative samples. $\theta$ is a set of matched (positive) pairs, the hardest negative sample is defined as the point closest to the positive pair in the $L_2$ standardized feature space, $L_2 = \sqrt{\sum_m (v_1 - v_2)^2}$, $m$ is output feature dimension by KP-FCNN. $v_1^i$ and $v_2^j$ are output features of the matched pair, $v^k$ is the hardest negative sample. $d(v_1^i, v_2^j)$ is the distance between $v_1^i$ and $v_2^j$. $c^{i,k} = \{1, \text{ if } g(x_1^i \neq x_2^k); \text{ else } 0\}$, where $g(\cdot)$ is the cluster obtained from k-means clustering in handcrafted geometric features space. $Q(v_1^j, v_2^k, c^{i,k})$ is the nearest distance between feature of a negative pair under the condition that $g(x_1^i) \neq g(x_2^k)$, $Q(v_1^k, v_2^j, c^{k,j})$ for the $x_1^k$ and $x_2^j$ similarly. $[x]_+ = \max(0, x)$, $|\theta|$ is number of the valid mined positive pairs, $|\theta_i|$ denotes number of the valid mined negative sample for $v_1^i$, $|\theta_j|$ denotes number of the valid mined negative sample for $v_2^j$. $t_p$ and $t_n$ are the margins of positive and negative pairs. Algorithm 1 describes our proposed auxiliary pre-task in detail.

**Algorithm 1** Algorithm of auxiliary pre-task

**Input**: Backbone architecture KP-FCNN; Dataset $X = \{x \in R^{N \times (3+d)}\}$;
**Output:** Pre-trained weights for NN.
**for** sampled mini-batch $X_1$ and $X_2$ from X until batch_limit **do**
   -From x, generate two regions $X_1$ and $X_2$ with a certain overlap.
   -Compute correspondence mapping (matches) $m$ between points in $X_1$ and $X_2$.
   -Sample two transformations: $T_1$ and $T_2$.
   -Compute geometric features.
   -Compute point clusters $k_1 = g(X_1)$ and $k_2 = g(X_2)$ by k-means.
   -Compute output features $V_1$ and $V_2 \in R^{n \times 64}$ by
$V_1 = f(T_1(X_1))$ and $V_2 = f(T_2(X_2))$.
   -Chose $v_1 = \{v_1^1, \cdots, v_1^i, \cdots v_1^p\}$ and $v_2 = \{v_2^1, \cdots, v_2^i, \cdots v_2^p\}$.
   -Backprop. To update KP-FCNN with contrastive loss $L_c(v_1, v_2)$ on the matched points.
**End**
**Return** network weights

## 3. Experiments

This section evaluates the proposed self-supervised learning method on several benchmark ALS point clouds. The proposed method is implemented on the PyTorch framework, on an Intel CoreTM i7-11700F CPU utilizing an NVIDIA RTX 3090 GPU with 24GB memory.

### 3.1. Experimental dataset

**Data for the Auxiliary task.** We considered the integrated scene categories and scales in the upstream auxiliary task and chose the DALES (Varney et al., 2020) point cloud dataset, which was collected using a Rigel Q1560 dual-channel system in British Columbia, Canada. The aircraft collection spanned an area of 330 $km^2$ and the entire data set is divided into 40 non-overlapping scenes, each of which covers 0.5 $km^2$ and more than 12 million points. The point cloud density is about 50 points/$m^2$ and the dataset includes eight object categories: ground, buildings, cars, trucks, telephone poles, wires, fences, and vegetation. The number of points in each category is presented in Fig. 6.

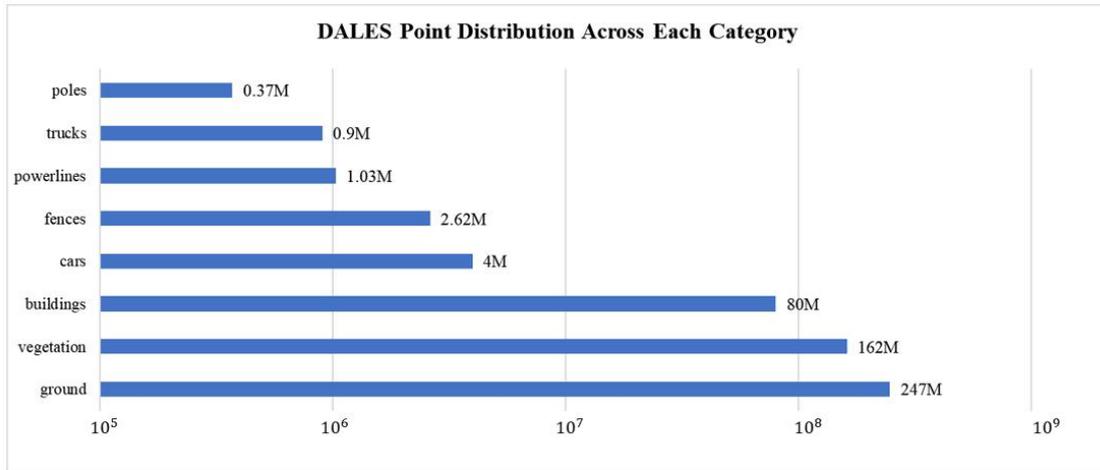

Fig. 6. DALES point distribution across object categories.

**Data for the semantic segmentation task.** In the downstream semantic segmentation task, we utilize the Vaihingen 3D semantic labeling benchmark dataset provided by ISPRS (Niemeyer et al., 2014), an ALS point cloud dataset captured from a Leica ALS50 system in Vaihingen, Germany. The average flight altitude is about 500 $m$, and the point density is about 6 points/$m^2$. Each point contains spatial coordinates (XYZ), intensity values, and the number of returns. The five attributes mentioned above are used as the network's input. The Vaihingen 3D dataset contains nine classes: power lines, low vegetation, impervious surfaces, cars, fences/hedges, roofs, facades, low shrubs, and trees. According to the official division, the Vaihingen dataset has two parts: the training set, an area of 399 $m \times 421\ m$ and the test set, an area of $389\ m \times 419\ m$.

In order to verify the effectiveness of the self-supervised model in the absence of labeled data and considering the distribution and quantity of each category, we exploit 10%, 20%, 40%, 60% and 100% of the training data to fine-tune the downstream semantic segmentation. The distribution of each category is presented in Table 2, and the point cloud area is depicted in Fig. 7 and Fig. 8.

Table 2. Data composition of the re-divided Vaihingen 3D dataset. Columns 2-6 are point number of nine categories in the five training sets, and the last column presents point number of nine categories in the test set.

| Class | 10% Training set | 20% Training set | 40% Training set | 60% Training set | 100% Training set | Test set |
|---|---|---|---|---|---|---|
| Powerline | 133 | 273 | 379 | 522 | 546 | 600 |
| Low Vegetation | 20,399 | 42,010 | 80,311 | 108,111 | 180,850 | 98,690 |
| Impervious Surfaces | 21,535 | 38,146 | 75,393 | 123,271 | 193,723 | 101,986 |
| Car | 636 | 919 | 1,847 | 2,311 | 4,614 | 3,708 |
| Fence/Hedge | 1,837 | 3,751 | 6,320 | 8,317 | 12,070 | 7,422 |
| Roof | 18,193 | 37,496 | 78,958 | 109,564 | 152,045 | 109,048 |
| Facade | 3,685 | 5,309 | 9,960 | 12,365 | 27,250 | 11,224 |
| Shrub | 7,895 | 12,695 | 23,874 | 31,296 | 47,605 | 24,818 |
| Tree | 11,569 | 16,663 | 32,458 | 55,941 | 135,173 | 54,226 |
| ∑ | 85,883 | 157,254 | 309,492 | 451,690 | 753,876 | 411,722 |

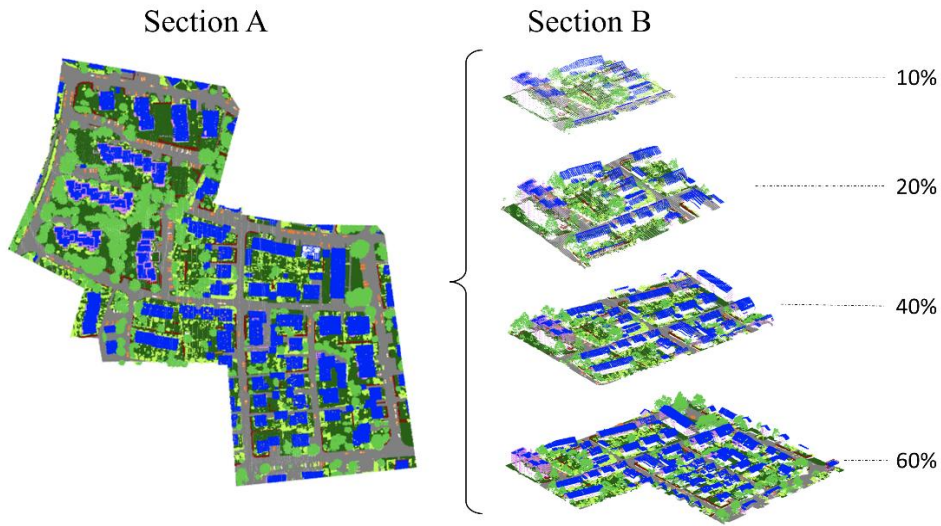

Fig. 7. ISPRS training set. Section A (left) is the full training set. Section B (right) shows the subsets of training data cropped from all training sets. The five subsets have the same category distribution.

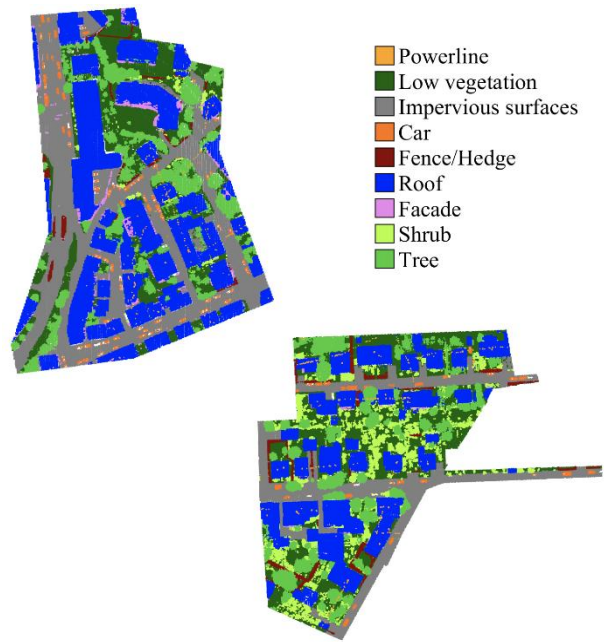

Fig. 8. ISPRS Vaihingen test set.

In addition to the ISPRS Vaihingen 3D data sets, we also utilize the LASDU dataset for verification, which was collected in the Heihe River Basin, Northwestern China, exploiting the Leica ALS70 system at a flying altitude of 1200 m. The average point density is about 3~4 points/m$^2$, the height difference range is about 70 m, and the area covers a city of 1 km$^2$. This dataset is divided into four regions, and the total number of points for all regions is 3.12 million. The LASDU datasets includes five classes: artifacts, buildings, ground, low vegetation, and trees. Each point contains its spatial coordinates (XYZ) and intensity value. Regions 2 and 3, illustrated in Fig. 9, are used as training set, while Regions 1 and 4 are used as test set. The training and testing set are presented in Table 3.

Table 3. LASDU dataset. Training and testing sets are divided according to the data provider's recommendations.

| Class | Training Set | Test Set |
|---|---|---|
| Ground | 704,425 | 637,257 |
| Buildings | 508,479 | 395,109 |
| Trees | 204,775 | 108,466 |
| Low vegetation | 210,495 | 192,051 |
| Artifacts | 66,738 | 53,061 |
| ∑ | 1,694,912 | 1,385,944 |

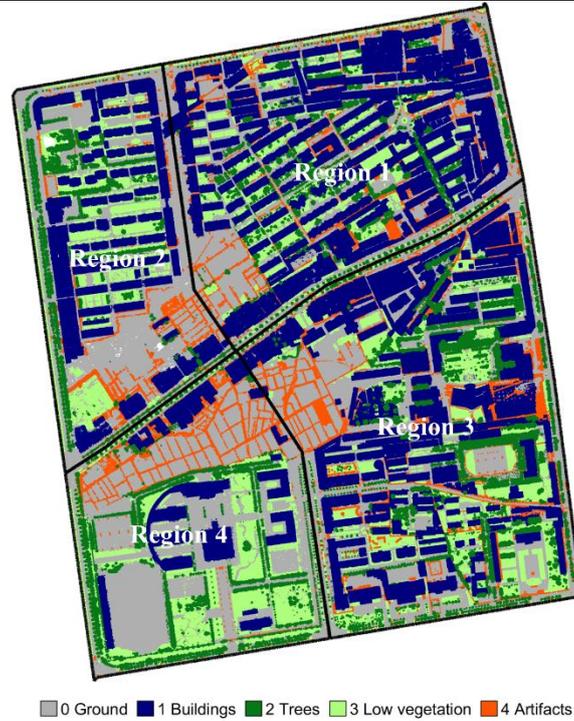

Fig. 9. LASDU data set. Annotated dataset with points of different labels rendered with different colors. The black border divides the entire area into four separated regions.

### 3.2. Evaluation metrics

We evaluate our method's segmentation performance utilizing the overall accuracy (OA) and average F1 score (Avg. F1). OA is the percentage of the correctly classified point out to the total points, and the F1 score is the harmonic average of the precision and recall of each category, defined as:

$$\begin{cases} \text{Precision} = TP/(TP+FP) \\ \text{Recall} = TP/(TP+FN) \\ \text{F1 Score} = 2 \times \dfrac{\text{Precision} \times \text{Recall}}{\text{Precision} + \text{Recall}} \end{cases} \qquad (4)$$

where TP, FP, and FN are the true positive, false positive, and false negative, respectively.

### 3.3. Parameters set up

**Parameters and preprocessing for the pre-training contrastive learning task.** The original DALES dataset contains 29 training and 11 testing tiles, and combined into 40 tiles for unsupervised contrastive learning. During data processing, we employed a grid sampling of 0.4 m to deal with extreme point cloud density variations. Moreover, from the entire point cloud, we randomly extract the vertices within a spherical volume, which are input to the network. We use the constant feature equal to 1, normalized intensity and normalized Z-coordinates within the sphere as input feature. The sphere's radius is 10 m, large enough to include several objects. The number of kernel points in the convolution operator is 19, properly set to balance computational cost and descriptive power. All five-layer networks involve a downsampling process, while the convolution radius of each layer and the downsampling grid size are presented in Table 4. During importing the two blocks for the contrastive learning process, the input point cloud is randomly rotated (0°~360°) and scaled (0.8~1.2) to augment it. For the hardest negative sample mining, we don't have to use all the points to mining, only a portion of the points are considered in the loss function (Choy et al., 2019a; Rao et al., 2020; Zhang et al., 2021). The parameter K for the k-means clustering algorithm is set to 9. For the contrastive learning loss, the number of positive pairs is set to 4096, the number of negative samples is set to 2048. Details about the quantity of positive and negative samples can be found in works (choy et al., 2019b). the threshold distance of the negative samples is set to 2.0, and for the positive samples is set to 0.2.

The stochastic gradient descent (SGD) optimization algorithm is employed to optimize the network weights, with a learning rate starting from 0.001. One epoch involves 8000 iterations, and we train the model for 50000 epochs.

Table 4. Convolution radius and down-sampling grid size of auxiliary pre-task five-layer Subsampling.

| Hyper-parameter | Layer1 | Layer2 | Layer3 | Layer4 | Layer5 |
|---|---|---|---|---|---|
| Down-sampling grid size(m) | 0.4 | 0.8 | 1.6 | 3.2 | 6.4 |
| Convolution radius(m) | 2.5 | 5.0 | 10 | 20 | 40 |

**Parameters and preprocessing for the downstream semantic segmentation task.** During dataset selection, datasets with insufficient data are selected on priority, thus the ISPRS and LASDU datasets are utilized to verify the gain afforded by self-supervision. During training data processing, we use grid sampling of 0.4 m, which is similar to the hyper-parameter setting process for the auxiliary tasks. For ISPRS datasets, we use the constant 1, intensity values, the number of returns and normalized Z-coordinates within the sphere as input 4D vector features. For LASDU datasets, we add intensity values and normalized Z-coordinates as additional feature to the constant 1, as input 3D vector. In the training stage, spheres are randomly selected into the network. In the testing stage, spheres are regularly selected, and each point is tested more than 20 times, on the one hand to ensure the integrity of the test set, on the other to obtain the average predicted value, similar to a voting scheme. As a result, the sphere's boundary points are well protected.

For the network parameters, the sphere's radius is set to 10 m, and the number of kernel points is set to 19. The convolution radius and the down-sampling grid size for each layer are presented in Table 4. The learning rate is set to 0.001, batch size to 4 and the decay rate is set to 0.98 at every 5 epochs. We train the model for 200 iterations in one epoch, and the convergence is expected after 50 epochs.

**3.3. Experimental results and analysis**

**3.3.1. Effectiveness of SSL**

We prove the efficiency of our self-supervised learning network on data comprehension. Therefore, we artificially tailor the ISPRS training scene randomly so that the amount of training data is only about 10% of the original training data. This strategy emphasizes the information gain provided by our self-supervised framework for downstream semantic segmentation tasks.

Table 5. Results with different methods for ISPRS datasets. Columns 3-11 show the per-class F1 scores, while the last two columns present the OA and Avg.F1 per method (all value are in %).

| Settings | Methods | Power | Low_veg | Imp_surf | Car | Fence/Hedge | Roof | Facade | Shrub | Tree | OA | Avg.F1 |
|---|---|---|---|---|---|---|---|---|---|---|---|---|
| training dataset (100%) | PointNet++ (Qi et al., 2017) | 57.9 | 79.6 | 90.6 | 66.1 | 31.5 | 91.6 | 54.3 | 41.6 | 77.0 | 81.2 | 65.6 |
| | PointSIFT (Jiang et al., 2018) | 55.7 | 80.7 | 90.9 | 77.8 | 30.5 | 92.5 | 56.9 | 44.4 | 79.6 | 82.2 | 67.7 |
| | D-FCN (Wen et al., 2020) | 70.4 | 80.2 | 91.4 | 78.1 | 37.0 | 93.0 | 60.5 | **46.0** | 79.4 | 82.2 | 70.7 |
| | RandLA-Net (Hu et al., 2020) | 76.4 | 80.2 | 91.7 | **78.4** | 37.4 | 94.2 | 60.1 | 45.2 | 79.9 | 82.8 | **71.5** |
| | DPE (Huang et al., 2020) | 68.1 | **86.5** | 99.3 | 75.2 | 19.5 | 91.1 | 44.2 | 39.4 | 72.6 | 83.2 | 66.2 |
| | KP-FCNN (Thomas et al., 2019) | 63.1 | 82.3 | 91.4 | 72.5 | 25.2 | **94.4** | 60.3 | 44.9 | 81.2 | 83.7 | 68.4 |
| | HAVANA (Ours) | 57.6 | 82.2 | 91.4 | 79.8 | 39.3 | 94.8 | 63.9 | 46.5 | 82.6 | 84.5 | 70.9 |
| training dataset (10%) | Pointnet++* | 58.6 | 66.2 | 78.6 | 28.9 | 25.1 | 87.1 | 61.2 | 43.1 | 72.6 | 72.1 | 57.9 |
| | KP-FCNN* | **63.2** | 76.4 | 85.9 | 50.4 | 18.8 | 84.7 | 54.7 | 40.8 | 69.9 | 75.9 | 60.5 |
| | HAVANA* (Ours) | 60.2 | **80.1** | 90.2 | 52.5 | 26.2 | 90.0 | 55.6 | 46.4 | 72.3 | 79.8 | 63.7 |

The segmentation results on the test data of the Vaihingen 3D dataset are reported in Table 5. We also compare our method with other published model, such as PointNet++ (Qi et al., 2017), PointSIFT (Jiang et al., 2018), KP-FCNN (Thomas et al., 2019), D-FCN (Wen et al., 2020), RandLA-Net (Hu et al., 2020) and DPE (Huang et al., 2020). We regard results of these methods as reference, and use KP-FCNN as the baseline model. The baseline networks are trained from scratch, while Table 5 presents the fine-tuned results of our HAVANA method using the pre-trained weights. It can be seen that our HAVANA trained with 100% training data achieves competitive improvement with fully-supervised models on the Vaihingen 3D test data. When reducing the training points to only 10%, the proposed method outperforms the baseline methods based on KP-FCNN, especially for small-numbered categories like cars and fences. When only 10% labeled training data are used for fine-tuning, our HAVANA method outperforms the baseline model, increasing OA up to 3.9% and Avg.F1 by 3.2%. The benefits of introducing unlabeled data to self-supervised learning are obvious. This proves that the proposed self-supervised methods effectively exploit the information learned from the auxiliary pre-tasks. It is important to note that improving the fully-supervised model with limited training data is worthwhile.

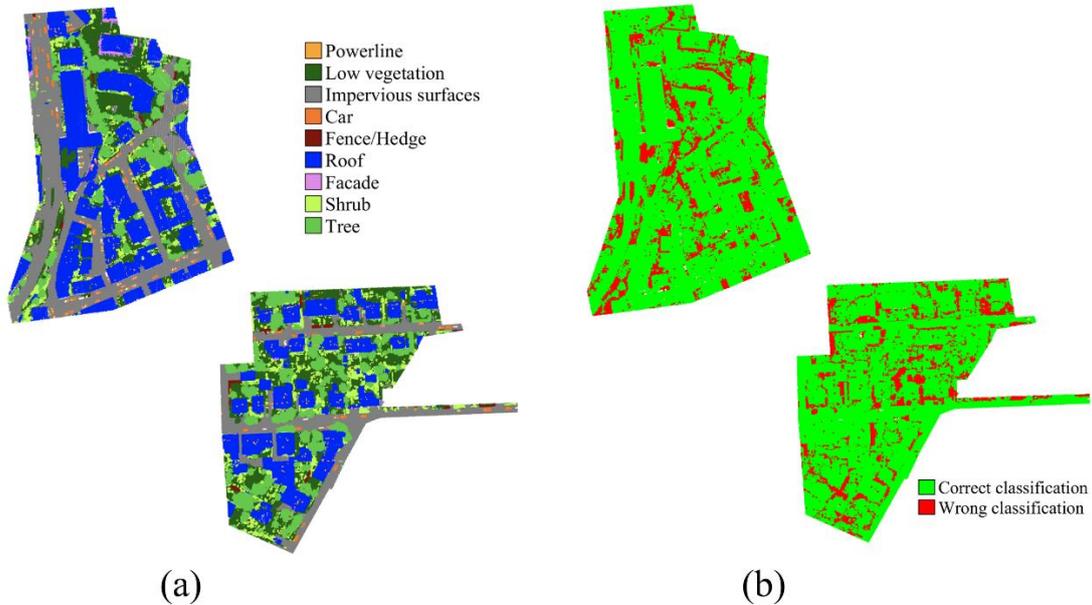

(a)                 (b)

Fig. 10. Classification results. (a) visualized classification results of our HAVANA on the ISPRS test set, (b) error map of our HAVANA on the Vaihingen 3D dataset.

Fig. 10 illustrates visualized classification results and the error map using our HAVANA method on the 100% Vaihingen 3D training dataset. The classification results show that HAVANA performs well in classification on the test set. From the error map, we can see that the labels of the majority of points can be predicted correctly. With reference to the visualized classification results, the majority of roof and impervious surfaces could be correctly classified. For the misclassified points, it is easy to find that facades are easily divided into roofs and fences into low vegetation. These categories indeed have similar geometric properties, local features are similar in neighborhood aggregation, and self-supervision improvement is also limited.

### 3.3.2. Hardest-Contrastive vs AbsPAN

In the subsequent trials, we compare the proposed HAVANA method against the Hardest-Contrastive strategy, a positive and negative sample selection strategy proposed in (Xie et al., 2020), with the corresponding results presented in Table 6. The originally reported Hardest-Contrastive strategy (Xie et al., 2020) utilizes the MinkowskiNet as the backbone network. Therefore, for a fair comparison, we replace the KP-FCNN with MinkowskiNet in our network, and presents a KP-FCNN variant with Hardest-Contrastive strategy.

For comparison, we only use 10% of the Vaihingen 3D training data for fine-tuning. By comparing MinkowskiNet and KP-FCNN, we find that KP-FCNN performs better on the proposed pipeline. The corresponding results reveal that if MinkowskiNet is used, the proposed AbsPAN presents improves OA and Avg.F1 by 0.4% and 1.9%, respectively, against the Hardest-Contrastive method. Under the KP-FCNN learning framework, the improvement is 0.9% and 1.0% for OA and Avg.F1, respectively.

Table 6. Effectiveness of the proposed AbsPAN strategy.

| Methods | OA (%) | Avg.F1 (%) |
|---|---|---|
| MinkowskiNet (Choy et al., 2019a) | 74.6 | 58.8 |
| MinkowskiNet (Hardest-Contrastive) | 76.4 | 59.5 |
| MinkowskiNet (AbsPAN) | **76.8** | **61.4** |
| KP-FCNN (Thomas et al., 2019) | 75.9 | 60.5 |
| KP-FCNN (Hardest-Contrastive) | 78.9 | 63.1 |
| KP-FCNN (AbsPAN) | **79.8** | **64.1** |

### 3.3.3. Performance of the SSL based on different backbone network

To see if the backbone network affects the proposed self-supervised pipeline, we utilize the self-supervised strategy with two other mainstream frameworks, PointNet++ (Qi et al., 2017) and MinkowskiNet (Choy et al., 2019a). The experiments consider the entire Vaihingen training data to fine-tune the network and evaluate its performance on the Vaihingen test data. We apply our SSL strategy to three frameworks respectively, and regard the three frameworks as reference methods. The corresponding results are presented in Table 7, revealing that the performance is improved regardless of the backbone network when the proposed pipeline is employed for semantic segmentation of ALS points. Regarding the three competing backbone networks, KP-FCNN performs best, and PointNet++ is the least appealing. When the KP-FCNN as the backbone network, Avg.F1 is improved by 2.5% compared to semantic segmentation methods without self-supervised learning.

Table 7. The effectiveness of self-supervision under three backbone networks.

| Methods | OA (%) | Avg.F1 (%) |
|---|---|---|
| PointNet++ (Qi et al., 2017) | 81.2 | 65.6 |
| PointNet++ (Our SSL) | **82.7** | **65.9** |
| MinkowskiNet (Choy et al., 2019a) | 82.9 | 66.2 |
| MinkowskiNet (Our SSL) | **83.7** | **67.8** |
| KP-FCNN (Thomas et al., 2019) | 83.7 | 68.4 |
| KP-FCNN (Our SSL) | **84.5** | **70.9** |

### 3.3.4. Performance with different amount of training data

In order to explore the ability of self-supervised learning to understand ALS point cloud scenes efficiently, we set up our contrastive learning framework (frame selection is biased towards lightweight model) to illustrate the adaptive changes of self-supervised learning under the five subsets of training data. Due to the uneven class distribution in the Vaihingen 3D dataset, as the area is too small to include all categories, we crop five subsets {10%, 20%, 40%, 60%, 100%} of the training area belonging to the official training set by ISPRS. The results are presented in Fig. 11 and Fig. 12. "Self-supervised Learining" denotes the result of fine-tuned with our pre-trained weights, "Train from scrach" denotes the result of train from scrach baseline.

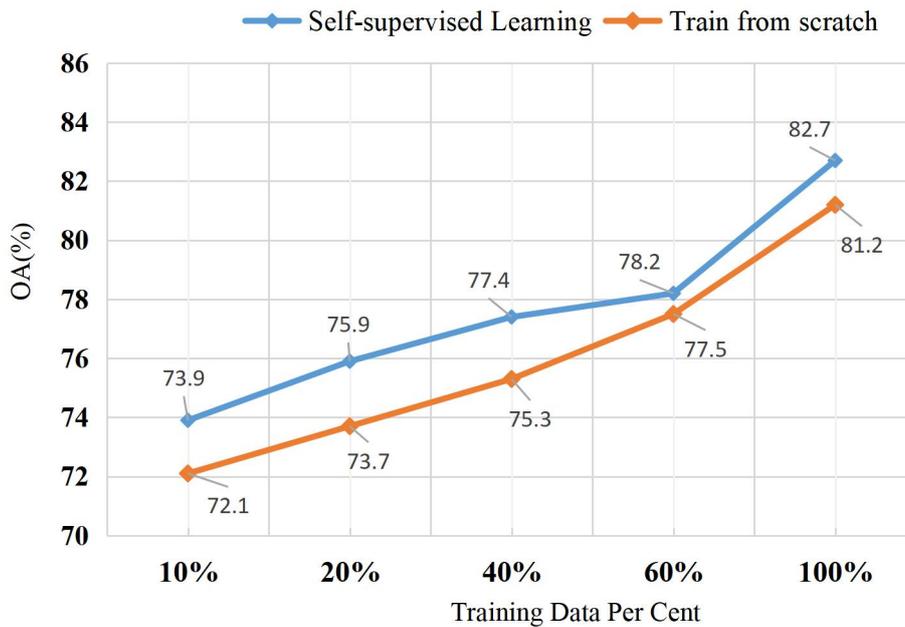

Fig. 11. Overall accuracy of different subsets in PoinNet++ framework.

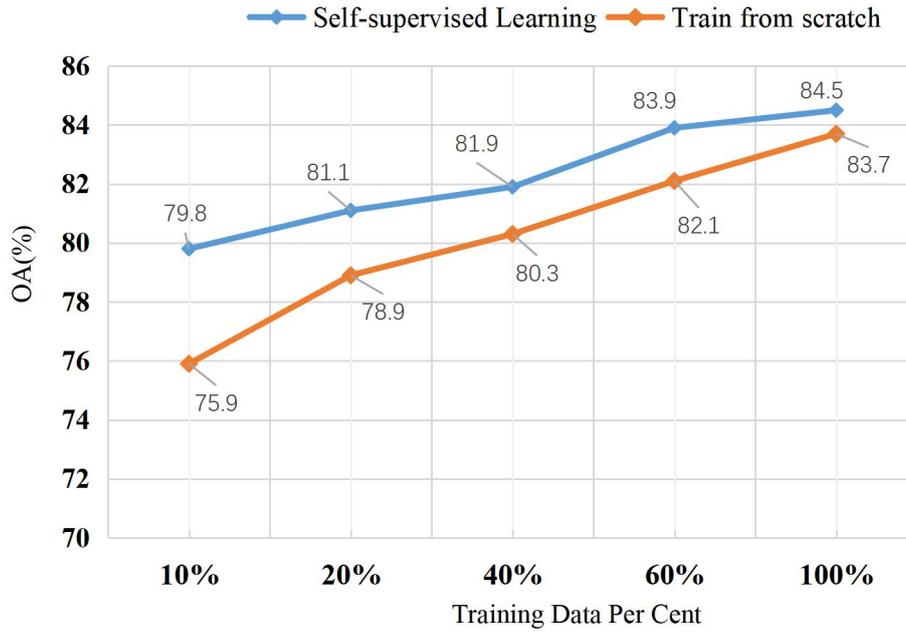

Fig. 12. Overall accuracy of different subsets in KP-FCNN framework.

For the experimental results, we take the average of the three experimental groups to reduce the impact of the variance. For the two baseline frameworks, we use the same experimental settings, and observe OA difference between fine-tuned with our pre-trained weights and trained from scratch under different training subsets, self-supervised pre-training weights are used for initialization methods have significantly improved the accuracy of semantic segmentation compared to training from scratch. From Fig. 11 and Fig 12, the trend is clear: the fewer semantic labels used for training, the greater the improvement in SSL. When the 10% labeled data is imported into the network, the OA improvement of the PointNet++ and KP-FCNN due to SSL is 1.8% and 3.9%, respectively. As the amount of semantic label data increases, the gap between self-supervision learning and direct training from scratch becomes smaller. It can be concluded that when labeled training data is insufficient, the proposed self-supervision learning is very important. However, the results show that using the SSL method under 10% of the training data still gives worse results than training from scratch using 100% of the training data. Thus if possible, labeling more samples is a good way to ensure the classification accuracy.

### 3.3.5. Further experiment results

In order to verify that the proposed HAVANA model is also improved on other ALS point clouds. The LASDU (Ye et al., 2020) dataset specifically designed for the semantic segmentation of ALS point clouds in very dense urban areas is chosen for further experiments. The proposed HAVANA method with KP-FCNN as the backbone is pre-trained on the DALES dataset, and then it is fine-tuned on the training data of the LASDU dataset. After that, the model is tested on the test data of the LASDU dataset. The results are shown in Table 8, where the results of the baseline methods are the ones reported in the related references, include Pointnet++ (Qi et al., 2017), PointSIFT (Jiang et al., 2018), KP-FCNN (Thomas et al., 2019), DPE (Huang et al., 2020), GraNet (Huang et al., 2021). These methods are all full-surpervised methods and training from scratch with LASDU official training.

Table 8. Comparison of classification results for LASDU datasets. Columns 2-5 show the per-class F1 scores, and the last two columns show each method's OA and Avg.F1 (all value are in %).

| Methods | Artifacts | Buildings | Ground | Low_veg | Trees | OA | Avg.F1 |
|---|---|---|---|---|---|---|---|
| Pointnet++ (Qi et al., 2017) | 31.3 | 90.6 | 87.7 | 63.2 | 82.0 | 82.8 | 71.0 |
| PointSIFT (Jiang et al., 2018) | 38.0 | 94.3 | 88.8 | 64.4 | 85.5 | 84.9 | 74.2 |
| KP-FCNN (Thomas et al., 2019) | **44.2** | 95.7 | 88.7 | **65.6** | 85.9 | 85.4 | **76.0** |
| DPE (Huang et al., 2020) | 36.9 | 93.2 | 88.7 | 65.2 | 82.2 | 84.4 | 73.3 |
| GraNet (Huang et al., 2021) | 42.4 | **95.8** | 89.9 | 64.7 | **86.1** | **86.2** | 75.8 |
| HAVANA (Ours) | **47.2(+3.0)** | **96.1(+0.3)** | **90.8(+0.9)** | **65.7(+0.1)** | **87.8(+1.7)** | **87.6(+1.4)** | **77.5(+1.5)** |

From the results, it can be found that the proposed self-supervised method performed best on the LASDU dataset, and for all five categories achieved the best performance. For the artifacts category with few training samples, the F1 score is improved by 3%, which benefits from the representation learning in the self-supervised auxiliary task. The OA of the HAVANA method reached 87.6%, and the Avg.F1 reached 77.5%, achieving the state-of-the-art performance.

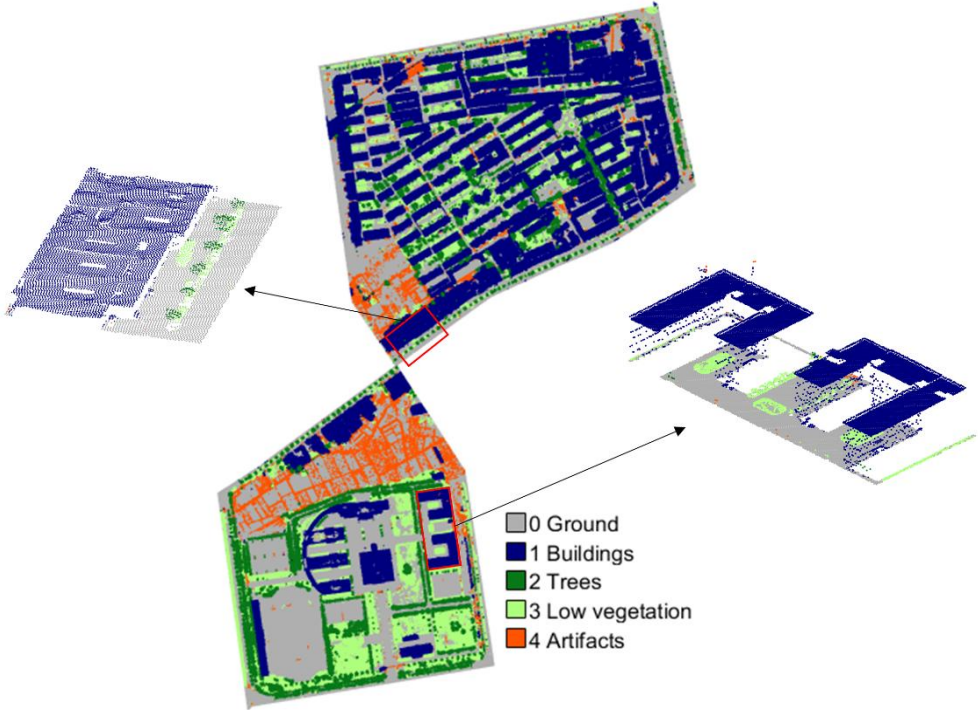

Fig. 13. Visualization of LASDU dataset classification results.

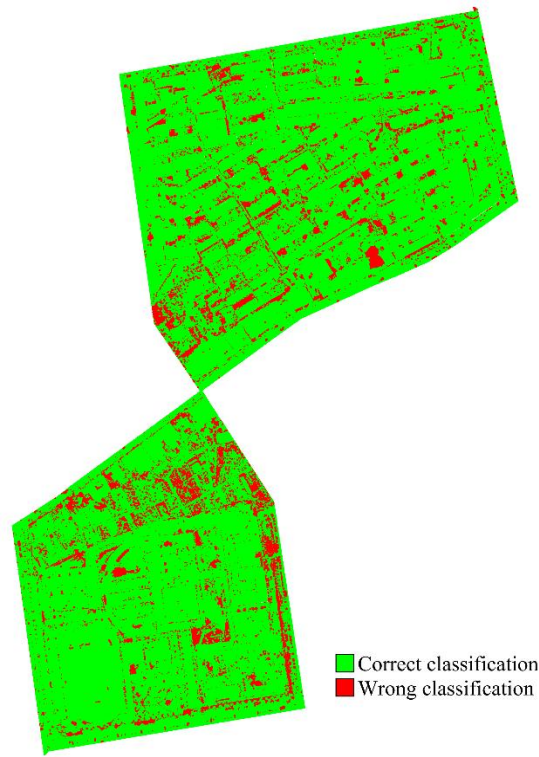

Fig. 14. Classification error map in LASDU dataset. The green and red points indicate the correct and wrong classification, respectively.

Fig. 13 visualizes the classification results of our HAVANA method on the LASDU dataset, highly realistic to the real urban scenery dataset. On LASDU dataset, our method achieves an appealing performance, highlighting the significance of our method. Specifically, the proposed scheme does not have obvious classification errors for high-density buildings, while for the ground vegetation and artifacts, it presents some minor misclassification. Although our method is appealing, more research is needed to improve its performance because classification currently relies primarily on spatial distribution information..Ground vegetation and artifacts,the latter two classes, are close in height and alternate in spatial distribution, imposing some errors. Looking at the classification error map of the LASDU dataset in Fig. 14, Most of the regions can be classified correctly, with a few partially incorrectly classification regions. With regard to Fig. 9, it is easy to find that the building's boundary classification is clearer, presenting obvious linear boundaries. This is a question worthy of further study.

### 3.3.6. Geometry Structure-Preserving on the Segmentation Details

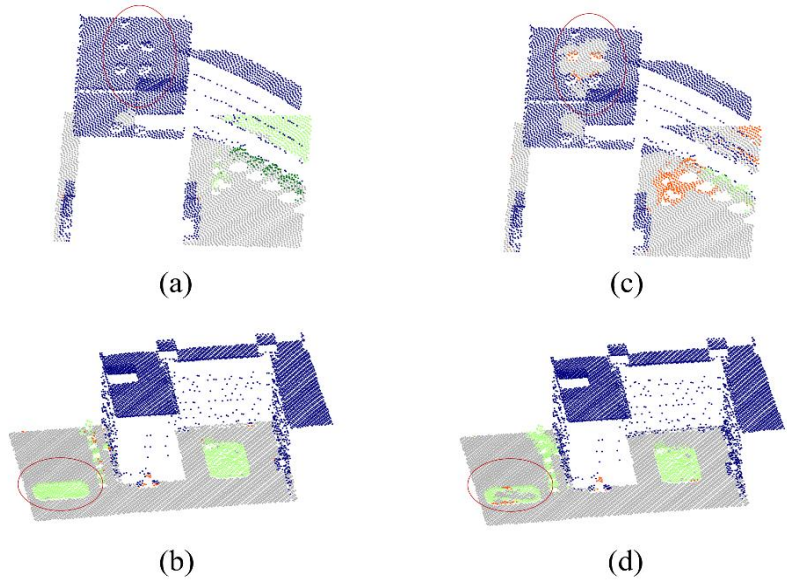

Fig. 15. Geometric structure of LASDU test sets. (a)-(b) SSL classification details, (c)-(d) classification details when training the network from scratch.

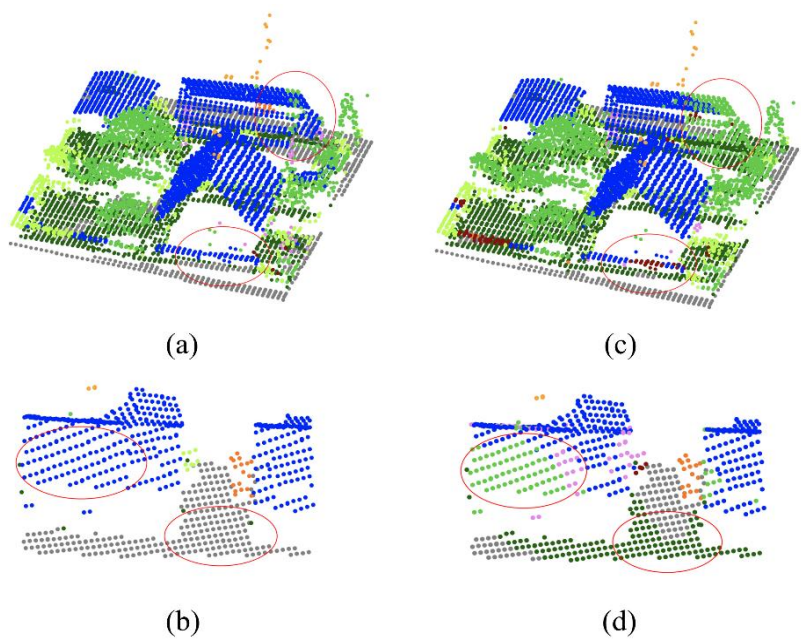

Fig. 16. Geometric structure of ISPRS test sets. (a)-(b) SSL classification details, (c)-(d) classification details when training the network from scratch.

By visualizing several experiment results, we find that the self-supervised-based pre-processing preserves the geometric structure information. It is often possible to improve the F1-scores of certain classes having different geometric properties (linearity and flatness), such as power lines, fences, roofs, and facades. Nevertheless, the F1 performance of other categories (cars and trees) does not improve significantly in Table 8. The results in Fig. 15 and Fig. 16 highlight the improvement of the geometric characteristics due to adopting a self-supervision scheme. Considering the LASDU test set, Fig. 15(c) shows that the buildings are mis-classified as ground, with the proposed method alleviating such errors (Fig. 15(a)). Low vegetation with distinct geometry structure can be correctly classified (Fig. 15(b)-(d)). Training the network from scratch commonly for geometric structure information is not sensitive, the structure information of point cloud can be transferred through self-supervised pre-training. Regarding the ISPRS test set, the geometric structure preservation due to SSL is evident in the junction area between two categories (Fig. 16(a)-(b)). Indeed, classifying houses and roads is more accurate under self-supervision learning. Considering training the network from scratch, in overlapping regions, two spatially similar categories are prone to error, e.g., areas where the tree covers a building (Fig. 16(c)-(d)). However, the supervision signal of the self-supervised auxiliary task guides the encoding of the entire neighborhood's geometric structure, making many geometric instances easier to distinguish. Our experiments highlight that the classification results of self-supervised categories have indeed improved, with obvious geometric features such as roofs and low vegetation.

## 4. Conclusions

This paper proposes a contrastive learning strategy for point clouds, a general unsupervised representation learning framework that performs iterative clustering and guides the feature learning direction. Through the ALS point cloud experiment shows SSL has great potential for point cloud semantic segmentation tasks when label is limited. The point cloud knowledge learned by our auxiliary task can be transferred to the downstream semantic classification task. Furthermore, the proposed method is easily integrated with a variety of deep learning frameworks. In the case of aerial laser point cloud with insufficient semantic labels, our experiments demonstrate the benefits of self-supervised contrastive learning. We conclude that self-supervision learning is clearly advantageous for semantic segmentation of limited labeled ALS point cloud.

There are also some limitations in the proposed method, KP-FCNN is a baseline network using geometric convolution to encode local point-wise features, which shall be improved to structure learning. The proposed method convinced that the better selecting negative samples are beneficial for the contrastive learning, it is worth investigate more robust sampling strategies for positive and negative samples in the future.

**Declaration of Competing Interest**

The authors declare that they have no known competing financial interests or personal relationships that could have appeared to influence the work reported in this paper.

**Acknowledgments**

This work was partially supported by National Natural Science Foundation of China under Grant 42171440, scientific research projects supported by the Department of education of Hunan Province under Grant 19K099, Hunan Province Key Laboratory of Key Technologies for Water Power Resource Development under Grant PKLHD201805 and

Hunan Provincial Innovation Foundation For Postgraduate under grant number QL20210060.

The authors thank ISPRS WG II/4 for organizing the Vaihingen 3D Semantic Labeling benchmark data and German Society for Photogrammetry, Remote Sensing and Geoinformation (DGPF) (https://ifpwww.ifp.uni-stuttgart.de/dgpf/DKEP-Allg.html) to kindly provide the Vaihingen 3D dataset (Cramer, 2010). The authors think College of Surveying and Geo-information, Tongji University for providing the LASDU dataset in this study (Ye et al., 2020). The authors think the University of Dayton for providing the DALES dataset in this study (Varney et al., 2020).